# Autotuning PID control using Actor-Critic Deep Reinforcement Learning

Vivien van Veldhuizen
11052929

Bachelor thesis
Credits: 18 EC

Bachelor *Kunstmatige Intelligentie*

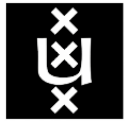

University of Amsterdam
Faculty of Science
Science Park 904
1098 XH Amsterdam

*Supervisor*
dr. E. Bruni

Institute for Logic, Language and Computation
Faculty of Science
University of Amsterdam
Science Park 907
1098 XG Amsterdam

Jun 26st, 2020

# Abstract

This thesis is an exploratory research concerned with determining in what way reinforcement learning can be used to predict optimal PID parameters for a robot designed for apple harvest. To study this, an algorithm called Advantage Actor Critic (A2C) is implemented on a simulated robot arm. The simulation primarily relies on the ROS framework. Experiments for tuning one actuator at a time and two actuators a a time are run, which both show that the model is able to predict PID gains that perform better than the set baseline. In addition, it is studied if the model is able to predict PID parameters based on where an apple is located. Initial tests show that the model is indeed able to adapt its predictions to apple locations, making it an adaptive controller.

# Acknowledgements

I would first like to thank my supervisor Elia Bruni, for always providing support and enthusiasm. Secondly, I would like to thank David Speck, for teaching us the much appreciated basics ROS. I also want to mention the support we have received over the ROS and HEBI forums, which has been a great help over the course of the project and finally, I would like to thank my colleague Leon Eshuijs, who made me strive for the very best within myself.

# Contents

# Chapter 1: Introduction

## 1.1 Problem Statement

Within the field of robotics, motor driven parts of a robot do not always react completely in tune with their control signals. In order to combat this problem, a control loop mechanism called Proportional Integral Derivative (PID) controller is often implemented. This controller uses feedback from the environment to continuously apply a correction based on the three hyperparameters: proportional, integral and derivative. The tuning of these parameters can get very complex, since the parameters are interdependent of each other. In addition, parameters cannot be further adjusted during robot operations after they have initially been tuned. This makes manually setting PID values a complex, time-consuming and oftentimes subjective task: values that are found often have satisfactory performance, but are not necessarily the most optimal. For this reason, this project will look into the possibility of using deep reinforcement learning to predict optimal PID values. The reinforcement learning algorithm will be designed to work on a robotic arm used for apple harvest, which must move towards and apple, pick it, move back and drop the apple in a harvesting basket. The goal for the applied learning algorithm is to be able to predict the PID parameters that are most optimal for every apple picking motion the robot executes. This thesis will therefore be concerned with determining in what way reinforcement learning can be used to predict optimal PID parameters for the apple harvest robot.

The remaining of this chapter will list some related studies that have also applied deep reinforcement learning for autotuning PID control. Subsequently, chapter 2 will provide the necessary theoretical background for this project, mainly focusing on the basics of PID control and reinforcement learning. Following this, chapter 3 will outline the design of the robotic arm, discussing its setup and how it is simulated. Then in chapter 4, the general methodology of the project will be given and the advantage actor-critic algorithm and learning environment will be defined. Chapter 5 follows up on the general method by specifying the technical details of the experiments that were conducted. Consequently, chapter 6 outlines the results generated in these experiments, which will be discussed and evaluated in chapter 7. The thesis concludes with chapter 8, which contains the conclusion, discussion and future works sections.

## 1.2 Related Works

Over the past years, some research into autotuning PID control using deep reinforcement learning has already been conducted. One study by Shi et al., 2018 used deep Q learning in combination with a multi agent system, which consists of three agents who all have one of the PID parameters to optimise. Because the MAS divided the problem into three separate spaces, the problem's complexity was reduced. However, convergence was not always achieved, since the three agents did not account for the other's actions.

Another study by Pongfai et al., 2020 also used deep Q learning, but combined it with a Swarm Learning Process. Results showed that the algorithm had better performance than other, more well known swarm learning and optimisation techniques, such as the whale optimisation algorithm and the improved particle swarm optimisation. A great drawback of using deep Q learning however, is that it can only be applied to problems with a discrete action space, as opposed to a continuous one. Should the algorithm need to predict a continuous numerical value, such as a PID gain, deep Q learning would not be a good fit.

An algorithm that might be used as an alternative to Q learning is the Deep Deterministic Policy Gradient (DDPG) algorithm, which was used for PID tuning in a study by Carlucho et al., 2020. Results showed a greater robustness compared to traditional PID controllers and the algorithm was more likely to perform better than advantage actor-critic in high dimensions. However, DDPG is an algorithm that explores much less than advantage actor-critic and for exploring a wide range of possible PID values, it might not always find an optimal solution.

Another alternative is a method that combines both elements of Q learning and continuous methods: Advantage Actor-Critic (A2C). One study by Mukhopadhyay et al., 2019 compared deep Q-learning and advantage actor-critic methods. They found that the deep Q agent could be adapted to work on a wider range of plant parameter variation. However, a robot optimised by the A2C agent showed better performance in trajectory tracking and was more precise, resulting in a better overall performance.

One of the earlier works that used actor-critic for autotuning PID parameters was conducted by Wang et al., 2007. Results showed that their PID controller, which used A2C to predict PID gains, produced more stable and robust results than ordinary controllers. A more recent study, conducted by Sun et al., 2019, also used an actor-critic method for optimising PID parameters, but instead of the regular advantage actor-critic, the asynchronous advantage actor-critic method (A3C) was used. This algorithm uses multiple agents that each update in accordance to an independent copy of the environment. While using the A3C algorithm resulted in less overshoot and a reduced steady state error, it was also a computationally heavy algorithm to use and since this project is a pilot study first and foremost, the added complexity of A3C might not be needed. The A2C algorithm however is more straightforward and also has not been researched in the context of autotuning PID control much over the past few years.

## 1.3 Research Question

This project will research how deep reinforcement learning can be used to predict optimal PID values for a robotic arm. It will do this trough the implementation of the A2C algorithm. Following from the related works, it is expected that the advantage actor-critic algorithm will be able to find gains that are optimal. It will be able to predict a full numerical range of PID values instead of being limited to a discrete action space like Q learning. Furthermore, A2C is also an exploratory algorithm, allowing the model to explore many possible PID values. Finally, A2C has been shown to result in good and robust performance, with precise trajectory tracking.

The aim of this project is primarily exploratory; it is a pilot study aimed at researching the possibilities for autotuning the particular apple harvest robot using deep reinforcement learning. In order to study this, we will first research whether A2C can be used to find PID gains that are optimal for one particular apple picking motion. Consequently, we will research if the A2C algorithm can be used to predict optimal PID values based on where the arm must move to.

## 1.4 Collaborations

The conducting of this research project relied on both a deep understanding of robotic simulation methods, as well as a deep understanding of advanced deep reinforcement learning techniques. Because of the complex nature of the project, it was set up to be a collaborative effort. My colleague Leon Eshuijs and I worked together intensively over the course of the project, first building the robot in a simulation environment and then collaborating on implementing the A2C algorithm on the environment we had created. While our theses are written completely independent of each other, including the interpretations of results, the overarching project setup was a joint effort.

# Chapter 2: Theoretical Background

This chapter will give an overview of the theory used in this project. Two sections will be discussed, each explaining concepts that are thought to be essential to this project. First, section 2.1 will discuss the basics of PID control. Subsequently, section 2.2 will give a short overview of general reinforcement learning and deep reinforcement learning, focusing on the difference between value-based and policy-based method and actor-critic which combines elements of both. This overview is not exhaustive by any means, but will be essential to understand the algorithm implemented in this project, which will be discussed later on.

## 2.1 PID control

A PID controller is one of the most widely used ways to regulate control processes. By continuously applying a correction to the output, a PID controller can help regulate for instance temperature in thermostats, speed in cruise controlled cars, or movement of a robotic arm. It does so by comparing the measured process value with a desired setpoint value. Subtracting the former from the latter results in the error. In order to minimise this error, a correction is applied based on Proportional, Integral or Derivative (PID).

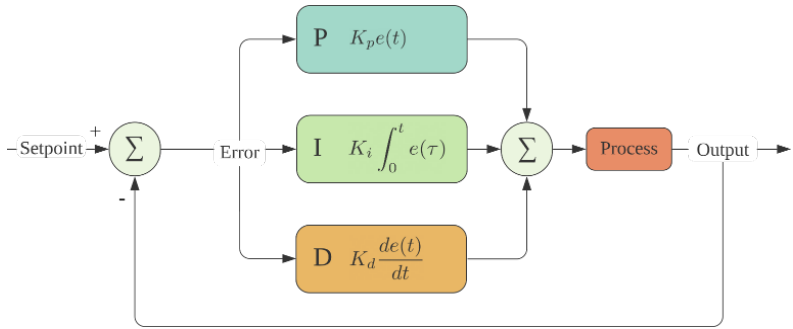

Figure 2.1: Diagram of a classic PID controller

Figure 2.1 shows a diagram of the PID control loop mechanism. As can be seen in the diagram, the process value is subtracted from the setpoint value, creating the present time error value $e(t)$. This error is then multiplied with the proportional, integral and derivative terms. In order to calculate these P, I and D terms, preset tuning constants $K$ are needed for every term. These gains are decided beforehand and essentially determine how much each of the PID terms has an influence on the total manipulated variable.

In the proportional term $P$, the error is multiplied by the gain $K_p$. In doing so, the manipulated output variable is set in linear proportion to the error. The proportional is a simple and efficient method for correcting the error, but a downside is that it can suffer from offset. This can for instance occur in a system with many weight changes, such as the current robotic apple harvest arm. When holding something of a certain weight, such as an apple, the arm's motors must change their output to suit that weight, as opposed to when it is not holding anything. The proportional cannot always correct such an error, because it only adjusts based on the present time and state of the robot arm.

The integral term $I$ can counter the offset problem. It also multiplies its gain $K_i$ with the error, but in this case, the error is integrated over time $\tau$, which goes from 0 to present time $t$. By integrating the error over time, the correction factor is adjusted based on how long and how far from the setpoint a certain error value is. The greater an error is or the longer it has persisted, the faster the integral will make the error go to zero. Because of this, the integral term can eventually cause the proportional's offset to be completely reduced. However, if the P and I term are quite aggressive and the error is brought down too fast, it can cause it to overshoot. The process value will then unwantedly go past the desired set point value.

This problem can be resolved by the derivative term. By taking the derivative of the error, the rate of change of the error becomes clear, which can then be brought to zero with the appropriate tuning constant $K_d$. The more rapidly the error approaches zero, or any other value, the greater the force applied by the derivative to slow it down. In this way, damping is added to the system and possible overshoot is reduced. Although this often causes the system to become more stable, a downside is that a greater derivative term makes the system more susceptible to noise.

Figure 2.1 shows the formulae for calculating each term of a PID controller. Once calculated, these terms are summed. This constitutes the manipulated variable which becomes the output of the control loop. Equation 2.1 shows the complete formula for PID control.

$$u(t) = K_p e(t) + K_i \int_0^t e(\tau)d\tau + K_d \frac{de(t)}{dt} \tag{2.1}$$

As mentioned earlier, the three PID terms are highly dependent on each other. Changing the value of one almost always has an impact on the other terms as well. Tuning the parameters is therefore a complex task. Current tuning techniques include trial and error, trajectory planning and the Ziegler-Nichols method, which sets gains from zero to a satisfactory value one by one (Graf, 2016). These methods are however still subjective and they are not guaranteed to find the optimal gains. This project therefore aims to find correct PID values by using deep reinforcement learning.

## 2.2 Reinforcement Learning

Reinforcement Learning (RL) is machine learning method that aims to to determine the most optimal sequence of actions to take in a particular environment. An agent is placed in an uncertain interactive, often game-like environment, and must find a sequence of actions trough trial and error which brings it closer to its goal. For every action taken, the agent receives feedback in the form of a reward or penalty. Since the agent's ultimate goal is to maximise the total cumulative reward, it can use this feedback to determine what actions to take.

### 2.2.1 Markov Decision Process

The reinforcement learning problem can be defined in terms of a Markov Decision Process or MDP (Sigaud & Buffet, 2013). In an MDP, at each instant in time, the agent is in its current state $s$ and can choose an action $a$ that is available for that current state. Transitioning from this state to a new state $s'$ is dependant on which action is taken. The probability for the agent to transition to a new state, given the current state and the chosen action is defined by the transition function $T(s'|s, a)$. Furthermore, the reward the agent is given is determined by the reward function function $R(s, a, s')$, which is the reward based on the current state $s$, the chosen action $a$ and the resulting state after taking that action $s'$. Note that this assumes that each state is only dependant on its previous state and not states that lie further in the past. This assumption is known as the Markov property. Assuming the Markov property greatly simplifies the problem of storing states in memory, as each state is only dependant on one other state at most. The ultimate goal of the problem is to find a function $\pi$ that maps a specific state to an action such that the cumulative reward is optimised. This function is called the *policy* and is defined by $a = \pi(s)$. In other words, the policy determines what action to take in a given state as to maximise the expected reward. Figure 2.2 gives a schematic overview of the reinforcement learning process in terms of a MDP.

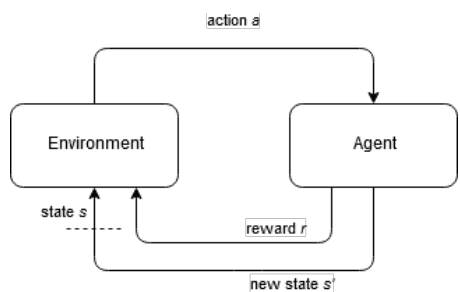

Figure 2.2: Interaction between agent and environment in a Markov decision process

### 2.2.2 Policy-based and Value-based Methods

In general, there are two main reinforcement learning methods for determining the optimal policy $\pi$. First of all there are policy-based methods, which learn or approximate the optimal policy function directly. Secondly, there are value-based methods, which first attempt to determine the optimal value function. The *value* is a measure of quality or usefulness and can be used to indicate how useful a certain state or action is with respect towards maximising the cumulative reward. By optimising the value function first, a policy can be subtracted that also has high value and can maximise the total reward efficiently. The algorithm used in this project, A2C, is from the actor-critic family, which combines elements of both policy-based and value-based reinforcement learning methods. Specifically, actor-critic methods are a type of policy gradient with added elements of temporal difference learning. These two elements will be briefly outlined in this section, before discussing actor-critic method in further detail. The section will assume some familiarity with basic reinforcement learning concepts, especially Q learning. However, should the reader like or require additional information on policy- and value-based reinforcement learning, they can consult appendix section A.1 and A.2.

**Policy Gradients**

One of the most well known types of policy-based reinforcement learning is policy gradients. As mentioned, the reinforcement learning problem encompasses how to determine an optimal policy such that the total cumulative reward of an agent is maximised. It is therefore essentially an optimisation problem. In policy gradients, to solve the optimisation problem, the policy is parametrised, allowing the optimisation problem to shift from the policy itself to the policy's parameters, which can then be optimised by gradient ascent (Sutton et al., 2000). For this problem, a parametrised stochastic policy is typically used, which is defined as $\pi_\theta(a|s)$, where

$\theta \in \mathbb{R}^d$ is the policy's parameter vector. The policy outputs a probability distribution of an action given a state $P(a|s)$ and therefore essentially determines how likely the agent is to chose a certain action in a given state.

The policy 's parameters can be optimised with respect to a policy objective function $J(\theta)$, which is defined in equation 2.2. Here, $t$ is the current time step and $s_t$ and $a_t$ are the state and action at time step $t$. The summation of all state-actions is also called the trajectory $\tau$. The objective function calculates the expectation $\mathbb{E}$ of the cumulative reward $r$ for a certain trajectory and thus essentially indicates how well the policy is performing.

$$J(\theta) = \mathbb{E}\Big[\sum_t r(s_t, a_t)\Big] \tag{2.2}$$

In order to optimise the policy's parameters with respect to the objective function, the gradient of the objective function $\nabla_\theta J(\theta)$ is needed. The gradient is the direction of the steepest change of the function and so, by using gradient ascent, the parameters vector $\theta$ can be moved closer to this gradient, therefore optimising the policy's parameters. These parameters are updated according to the gradient update rule, which is given in equation 2.3.

$$\theta_{t+1} = \theta_t + \alpha \nabla_\theta J(\theta_t) \tag{2.3}$$

where the objective gradient is given by

$$\nabla_\theta J(\theta) = \mathbb{E}\Big[\Big(\sum_t \nabla_\theta \log \pi_\theta(a_t \mid s_t)\Big)\Big(\sum_t r(s_t, a_t)\Big)\Big] \tag{2.4}$$

Here, the parameter vector $\theta$ for the next time step $t+1$ is given by the parameter vector of the current time step $t$ plus the gradient of the objective function multiplied by the learning rate $\alpha$.

**Temporal Difference Learning**

Temporal Difference (TD) learning is an approach to value-based reinforcement learning, of which Q-learning is one of the most well-known examples. Because it is a value-based method, TD-learning is concerned with estimating a value-function which can be used to determine the optimal states and actions for a particular policy. The value-function of a state $s$ is given by $V^\pi(s)$, where the value $V$ is the expected future reward that an agent will receive assuming that it is in a certain state $s$ and continues with the current policy $\pi$ thereafter. Unlike other value-based techniques such as Monte Carlo methods, TD-learning does not wait until it has received the total reward at the end of an episode to determine the value function $V$ (Sutton, 1988). Instead, it estimates the value-function after every few steps, making an estimation of the final reward and then using this estimation to update the state's value. The using of estimations to update other estimations is called *bootstrapping* and it is often seen to reduce variance and speed up learning when compared to non-bootstrapping methods such as Monte Carlo (Sutton & Barto, 2011).

As mentioned, in TD-learning a new estimation can be made after $n$ number of steps. One of the most simple and well-known TD-learning methods is one-step TD or TD(0), where the value function is updated after one step. This is defined mathematically as

$$V(s_t) \leftarrow V(s_t) + \alpha\Big(R_{t+1} + \gamma V(s_{t+1}) - V(s_t)\Big) \tag{2.5}$$

Here, the value $V$ of state $s$ at time step $t$ is updated based on the error between the estimated value and the actual returned reward, also called the *TD error*. In order to calculate this error, the estimated value of state $s_t$ is subtracted from the actual reward received from the state in the next time step $t+1$ and the *discounted* expected value of the next state. The discount factor $\gamma$ determines the importance immediate rewards have in regards to rewards that are farther in the future. The TD error is then multiplied by the learning rate $\alpha$, which determines how much influence the error has on the total update, after which the whole term is added to the value estimate of the current state. By continuously comparing the new reward to the value estimate and updating the prior estimate based on this error, the value function can be approximated better with every step.

### 2.2.3 Hybrid Methods: Actor-Critic

Temporal difference methods, such as Q-learning, and policy gradients are both widely used reinforcement learning techniques and both come with it's own advantages and disadvantages. Value-based methods often suffer from instability and poor convergence, because they do not learn the optimal policy directly, but only extract it from an approximated optimal value function. Policy-based methods such as policy gradients are thus generally more stable and converge more often. They can also be stochastic, meaning that the policy gives a probability distribution over the possible actions instead of one discrete action. This also makes policy-based methods more effective in continuous and high dimensional spaces. However, a TD-learning algorithm that does converge is often computationally more efficient. In contrast, policy gradients generally learn slower and often suffer from high variance in their estimates (Sutton & Barto, 2011).

Actor-Critic (AC) methods are a type of reinforcement learning that attempt to combine the advantages of both value-based and policy-based methods (Barto et al., 1983). The core of actor-critic is that there are two models: one that approximates the policy, the *actor*, and one that approximates the value function, the *critic*. The actor resembles a policy gradient model as defined above: it uses a neural network to approximate a policy function by optimising its parameters. This network in turn indicates how likely a certain action is to have high reward. Actor-critic differs from a regular policy gradient in that it does not use the total reward to compute the objective gradient, as was the case in equation 2.4. Using the total reward would mean that updates can only be performed at the end of an episode, because it is only then that the total reward is known. Instead, in actor-critic the total reward $\sum_t r(s_t, a_t)$ is replaced with a value function that approximates the maximum of the expected future reward. This value function is approximated by the critic. By using a value function instead of the total reward, it becomes possible to use bootstrapping, meaning that the expected reward can be determined for individual steps instead of only at the end of an episode. The actor-critic gradient is thus defined as

$$\nabla_\theta J(\theta) = \mathbb{E}\Big[\Big(\sum_t \nabla_\theta \log \pi_\theta(a_t \mid s_t) V_w(s_t, a_t)\Big)\Big] \qquad (2.6)$$

where $\theta$ is the parameter vector for the actor model, which approximates a policy function $\pi_\theta$ and $w$ is the parameter vector for the critic model, which approximates any value function $V_w$. Multiple variants of the value function can be used in actor-critic, but by far the most adopted is the *advantage function* (Schulman et al., 2015). The advantage function is also used in this particular project, so its benefits will be discussed in further detail in section 4.3. Figure 2.3 shows a simplified diagram of the actor-critic process. Trough the use of bootstrapping, the actor and critic alike can update their model weights for every step in the episode, without having to wait until the end of the episode to perform one update. This reduces the high variance that actor-only policy gradients suffer from and greatly speeds up the learning process (Baird & Moore, 1999).

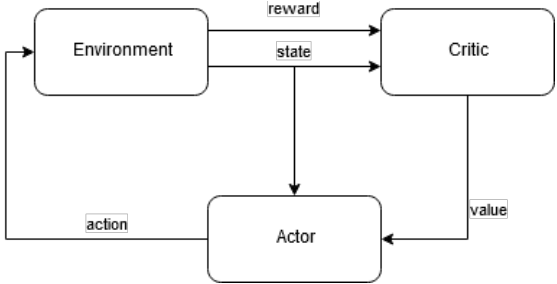

Figure 2.3: Basic actor-critic process

# Chapter 3: Robot Design

While the autotuning of PID control described in this project will ultimately be implemented on a real robot, the training and testing of the algorithm was conducted on a simulated version. This made it possible to safely explore multiple implementation options, without risking breaking the arm. In this chapter, the general setup of the simulated robot will be discussed. First of all the technical details of the robot build will be quickly mentioned. Consequently, the main three components of the robotic simulation framework will be discussed, which consist of the general simulation framework trough *ROS* and *Gazebo*, the PID control trough the *ROS Control* package and the motion execution framework *MoveIt!*. The chapter will conclude with an overview of how these three components integrate and work together.

## 3.1 Robot Setup

The robot arm used in this project is a two-link arm, mounted on a rail and with a gripper attached as end effector. The gripper and arm were not attached at the beginning of this project and had to be joined together in a simulation, so that the robot could be controlled and tested properly. Figure 3.1 shows a simulation of the joined robot arm. The arm is connected to the rail via a prismatic joint, which allows the arm to turn a full circle around the rail. The two parts of the arm are connected by a revolute elbow joint, which is set to turn 180 degrees so as to not cause collisions between the arm an rails. Furthermore, a gripper is attached to the arm by a revolute wrist joint. In addition to being controlled horizontally by the arm's wrist joint, the gripper can be rotated around its axis and the left and right gripper parts can both move separately, each with their own revolute joint. When fully stretched, the arm has a length of 1 meter and the rails the arm is mounted on is 1.5 meters tall.

Each of these joints is driven by a motor, which also means that each of these actuators has their own PID-controller. The two motors in the elbow and shoulder part of the arm are supplied by HEBI Robotics[1], a company that supplies robotic hardware and software tools aimed at simplifying the building of functional robots. The remaining actuators were supplied by the Dynamixel[2] brand. The two actuators supplied HEBI Robotics were controlled via a non-standard HEBI framework, as opposed to the other actuators which were controlled trough a general ROS architecture.

## 3.2 Control Components

### 3.2.1 ROS

The simulating of the robotic arm is done trough ROS[3], or Robotic Operation System, a framework that provides the tools needed to help create robot applications. In this project's particular instance, the main purpose of ROS is to visualise and control the robot in a simulation environment. One of the key features that enables ROS to do this is a file format called Unified Robot Description Format (URDF). In this URDF file, the main components of the robot can be defined, such as the links, joints and actuators. This allows the robot to be simulated in an environment such as rviz[4], ROS's own visualisation tool, or an external simulation tool. For this project, the robot was simulated in Gazebo[5], a simulation environment that also includes a simulation of physics. Since physics are an important part of why PID-controllers are needed, it is important to use a simulator that can mimic real physical conditions as closely as possible.

---

[1]HEBI Robotics, https://www.hebirobotics.com/
[2]Dynamixel Robotics: http://www.robotis.us/dynamixel/
[3]ROS: https://www.ros.org/
[4]RViz: http://wiki.ros.org/rviz

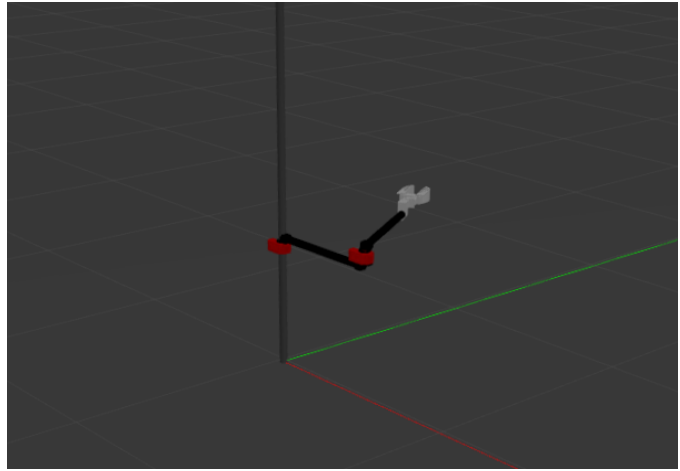

Figure 3.1: The robot arm with attached gripper simulated in Gazebo

Another key feature of ROS is to facilitate communication between the simulated robot and other written code. In ROS, each executable code process is represented by a node. These nodes all represent a computational task, such as observing the gains of a certain actuator, sending that information to another node which can manipulate the gains, or a node that can move the robot into a new position. Such communication is done by sending and receiving messages, which are published on topics. These topics can be seen as certain categories on which information is transmitted. A node can subscribe to topics to receive information from that category, or can publish messages which tell other nodes information on a certain topic.

### 3.2.2 PID Controllers

In order to build and control a PID controller for the robot, an external ROS package called `ros_control`[6] was used. The `ros_control` package contains multiple controllers and the appropriate interfaces to interact with them trough the ROS framework. For this project, a general controller called `joint_trajectory_controller` was used, which could apply PID control trough three types of input: position, velocity and effort. For each of these interface types, the controller would receive input from the joint state data of the robot's encoders, apply the PID control loop, convert the output to effort -which in this particular instance is the amount of current that is supplied to the motor- and feed it to the robot actuators. For the Dynamixel actuators, this project used the default input type set by MoveIt, which was effort. Effort allowed for a direct, low-level type of control, as input was directly applied as current to the actuators[7].

---

[5]Gazebo Simulator: http://gazebosim.org/
[6]http://wiki.ros.org/ros_control
[7]ROS Control Types: https://www.rosroboticslearning.com/ros-control

**HEBI PID Control**

Because the HEBI actuators were controlled by a framework separate from general ROS, the PID control worked slightly different as well. Instead of providing input either trough position, velocity or effort, these three types were combined into one architecture[8]. This architecture is shown in figure 3.2.

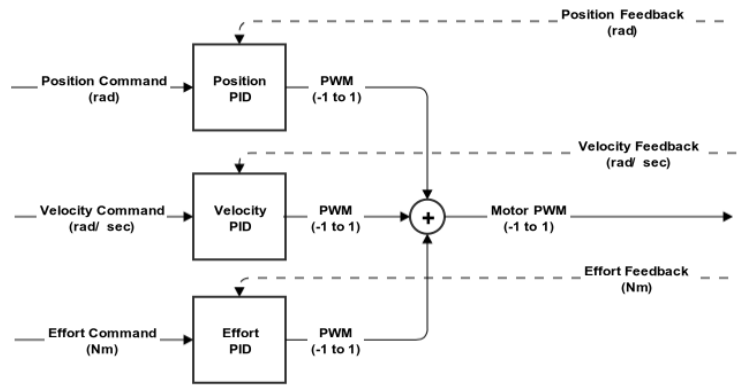

Figure 3.2: HEBI Control Strategy

As can be seen in the diagram, the three input types are combined into one and summed to generate a motor current output. In order to make the HEBI actuator's control more similar to that of the other non-HEBI actuators, all but one of the three control types were set to zero. In this way, results between actuators would be more general. The input control that was chosen for HEBI was position, as opposed to the other controllers, which used effort. Early experiments showed that for HEBI, position control was much more precise, with velocity and effort often resulting in unstable movements. In addition, the HEBI website recommended to use either position or velocity, but preferably not effort, as it could lead to instability[9].

### 3.2.3 MoveIt Motion Planning

In order to train the model and test PID gains, the robot arm does not only need to be visualised and equipped with a PID controller, it also needs to be able to execute movements. In particular, it needs to be able to move towards an apple somewhere in the workspace, grasp it and pick it with the gripper and then move back to put the apple in a bin or basket. This means that, based on the coordinates of the apple, the robot needs to plan and follow a trajectory in order to grasp it. In the simulation environment, this motion planning is accomplished trough *MoveIt Motion Planning Framework*[10]. MoveIt is a framework that plans a motion of the robot based on its start state and desired end state. Before MoveIt can be used on a custom robot, it needs to be configured properly. In order to generate the needed configuration files, a tool called *MoveIt! Setup Assistant*[11] was used. This setup wizard can generate all required configuration files based on information given by the user via a graphical user interface. The basis for the configuration is formed by the URDF file, which serves as the input for the setup assistant. Using the information about the robot's components, the correct planning groups can be defined. These planning groups consist of all the robot components that need to make up one independent motion together. In this case, these movements are the reaching from or towards the apple of the robotic arm and the picking of the apple by the arm's gripper. There are therefore two planning groups: the arm and the gripper. The MoveIt! Setup Assistant can also be used with both Gazebo and `ros_control`. For the first, an adjusted URDF file is generated so that the robot can move in Gazebo. For use with `ros_control`, the setup assistant generates the needed controllers to actuate the robot joints.

---

[8] HEBI Control Strategies: https://docs.hebi.us/core_concepts.html#control-strategies
[9] HEBI Gain Tuning Recommendations: https://docs.hebi.us/core_concepts.html#controller_gains
[10] https://moveit.ros.org/
[11] MoveIt! Setup Assistant Documentation: http://docs.ros.org/melodic/api/moveit_tutorials/html/doc/setup_assistant/setup_assistant_tutorial.html?highlight=assistent

After setting up the MoveIt interface, the robot can move by specifying a desired goal state. While this can also be set trough a graphical user interface, this project uses Python code to specify desired end states, because the actor-critic algorithm needs to be able to command the movement of the robot arm. This functionality is provided by *MoveIt Commander*[12]. Using the MoveIt Commander, it becomes possible to give the robot arm desired coordinates in joint space, or specify a certain end effector pose to move to. The planning of this motion from the initial state of the robot arm to the desired goal state consists of two main steps: planning a path and executing a trajectory.

**Paths and Trajectories**

A path in MoveIt is defined as a set of joint coordinates, or waypoints, that specify the sequential positions the robot must reach to ultimately reach the goal state. Paths are generated by a motion planner, which in this case is a planner called *OMPL*[13]. The OMPL planner generates paths in a purely kinematic way; it compares the desired goal state to the current robot state and specifies a kinematic path to follow to that goal. This is called planning. A trajectory on the other hand, is a path with velocities and accelerations added to it. Creating trajectories is done by a separate trajectory API, which in this case is one supplied by HEBI Robotics[14]. The trajectory generator uses time parameterisation to assign time points to the different waypoints, which each specify by what time the robot must have reached a particular waypoint. In this the generator accounts for joint and acceleration limits specified by the user in one of the control files. Based on these time points, the required velocities and accelerations between each subsequent pair of waypoints can be computed and this data can be sent to the controllers and actuators.

## 3.3 Integration of Control Components

Figure 3.3 shows a simplified overview of the major setup components mentioned above. The bottom layer of the diagram forms the robot layer, which is simulated trough Gazebo. Secondly, the middle layer forms the control layer, with the PID controller implemented trough the control package. Finally, the top layer is formed by MoveIt and all its related components. The centre of MoveIt control is formed by the `move_group` node. This node works as an integrator for all the different smaller components, receiving the URDF file of the robot and all the configuration files generated by the MoveIt Setup Assistant. The move group first receives a request for a motion planning service trough the MoveIt Commander. For example, it might be running a Python script whose objective is to move to a point $[x, y, z]$. Both this request and the current joint states of the robot are received in the move group node and forwarded to the kinematic motion planner OMPL, where the current state and desired goal state are converted into a path. The path then goes into post-processing and is converted into a trajectory by the HEBI Trajectory API, after which it goes back to the move group node. Using the `FollowJointTrajectory` action service, the generated trajectory is executed and reaches the controller command interface. This controller simultaneously receives information on the current joint states from the robot encoders. Using this information, the PID controller calculates the error between the commanded states and the actual states and applies a correction to the effort, which is supplied to the robot actuators. The controller does this for every step in the trajectory, thus continuously applying a correction while moving.

---
[12]MoveIt Commander: http://wiki.ros.org/moveit_commander
[13]OMPL Motion Planner:http://ompl.kavrakilab.org/
[14]Hebi Trajectory API: https://docs.hebi.us/core_concepts.html#trajectory_overview

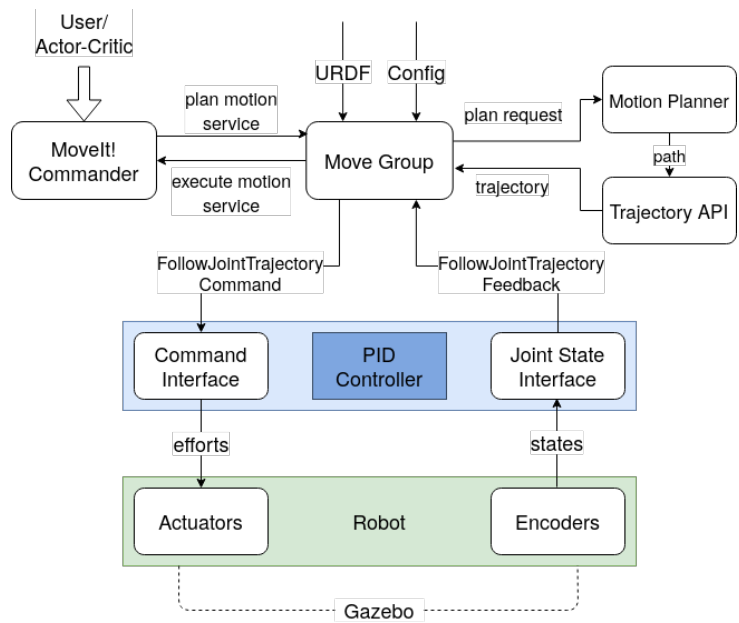

Figure 3.3: Integration of Moveit!, ros_control and Gazebo

# Chapter 4:   Methodology

Consider an apple at a certain coordinate in Cartesian space that the robot arm has to pick. It is the robot's objective to move towards this apple, pick it and then move it back to the starting point as quickly as possible. In order to complete this task, the robot has been given control over the PID gains of each of its actuators. Trough the use of an actor-critic algorithm, it must determine the optimal PID gains for each individual apple picking motion that it executes, making it as fast as possible. This chapter will describe the general method of how this project aims to solve the problem defined above. To begin, a short overview of initial testing and motivation for choosing the advantage actor-critic algorithm will be given. Secondly, the environment for the reinforcement learning problem will be defined. Then follows a more in depth definition of this employed algorithm and finally, the implementation of this particular algorithm on the environment and robot will be outlined.

## 4.1   Preliminary Experiments

This project implements a one step, continuous version of the Advantage Actor-Critic algorithm (A2C) to solve the problem at hand. Before arriving at this algorithm however, some other options were discussed and briefly tested. In order to quickly implement and compare multiple techniques, preliminary tests were conducted on two environments that were more simple than that of the robot simulation. These environments were supplied by OpenAI Gym[1]. The OpenAI Gym environments have the benefit of being very easy to implement, allowing the user to only have to focus on supplying the algorithm to use on the environment. Additionally, a great number of reinforcement learning algorithms for Gym's environments is already available open source, making it easier to compare multiple implementations. For the preliminary testing, two Gym environments were employed: the cart pole balance problem[2] and the slightly more complex lunar lander problem[3]. The cart pole balance problem is a classic reinforcement learning problem where a cart must be pushed left or right in order to balance a pole standing on top of it. It is a relatively simple problem that is often used in reinforcement learning research for the evaluation of algorithms (Shi et al., 2018, Sun et al., 2019). In the lunar lander problem, an air vehicle must land within a bounded landing region by continuously applying control, either in the form of four discrete actions, or by providing continuous values of orientation and velocity. The lunar lander problem is similar to the cart pole problem, but due to its larger, possibly continuous action space, it more closely resembles the current robot simulation environment. A PID controller was added to both of these environments: the reinforcement learning model had direct control over the PID controller, which in turn controlled the motors of the cart or lunar lander.

As discussed in the previous section, currently the action space of the problem is considered to be continuous; the agent has to predict real numerical values. This was not always the case however, as the project started out by defining a discrete action space. Instead of directly predicting a number, the agent had to choose one of the three PID gains at every step to add or subtract a predefined value from. Two reinforcement learning techniques were tested using this setup: deep Q-learning and advantage actor-critic. The results of these initial tests on the Gym environments indicated that the discrete action space was insufficient for finding well performing PID gains, as both algorithms showed no improvement over episodes and did not find gains that performed better than the default settings. The decision was therefore made to make the action space continuous, so as to not limit the extend to which the PID gains could be adjusted per episode. This also meant that deep Q-learning could no longer be used, as it only works on discrete action spaces. Therefore only A2C was tested on the continuous version of the PID lunar lander problem, where it showed an increasing reward over time. This motivated the choice to continue working with a continuous action space and the A2C algorithm for the real robot simulation.

---

[1] OpenAI Gym: http://gym.openai.com/
[2] OpenAI gym CartPole environment: https://gym.openai.com/envs/CartPole-v0/
[3] OpenAI gym LunarLander environment: https://gym.openai.com/envs/LunarLander-v2/

Another reason for choosing to use the A2C algorithm for this project is its potential to be extended for use with a more adaptive PID controller. Currently, the agent chooses PID gains and then executes a full motion towards the apple and back. In the future however, a likely research direction is to study the possibility of updating the PID gains at multiple time steps within the full motion, for example before and after holding the apple. In this situation a regular policy gradient might suffer from high variance due to a delayed reward over steps, but the A2C algorithm, which uses the advantage function and bootstrapping, would fit such a problem well. The current implementation of A2C is a one-step version, but it might easily be adapted to work with multiple steps, should that be required one day.

## 4.2 Environment

Before detailing the implementation of the A2C algorithm, we must first define the reinforcement learning problem in terms of an agent interacting with an environment. The agent in this problem is the robot arm. It is equipped with the ability to move towards a certain coordinate, utilising a controller that needs to be set with PID gains. These gains are decided by the A2C algorithm; the agent consists of an actor which determines which actions to take and a critic which attributes values to these decisions. For the environment, the most important components that need to to be defined are listed below.

**State Space**

A state in this environment is set of coordinates, consisting of an $x$, $y$ and $z$ direction. The agent starts each episode at the harvesting basket and can move to any apple coordinate $[x, y, z]$ from there, where it will pick the apple and return to the home state. The state space is defined by the workspace of the robot: any apple coordinate that falls within the working space of the robot arm can be a valid state. The exact details of the work space will be defined in chapter 5.

**Action Space**

Moving between the start state and the apple coordinate state can be done by planning a path and consequently executing the trajectory, utilising a trajectory controller that requires a set of three PID gains for each actuator: $K_p, K_i$ and $K_d$. These gains need to be decided by the agent, meaning that actions in this problem are defined by the setting of gains. Because PID gains can be set to a continuous range of numbers, the action space in this environment is continuous as well. However, since a PID controller is essentially a negative feedback control loop, negative gains would result in positive feedback, potentially making the system unstable (Graf, 2016). The action space will therefore be defined to only include positive real numbers. Moreover, each individual PID gain generally performs best within a certain range of values: $K_p$ could vary between 0 and 1000, while $K_i$ often performs best within a much smaller range, for example from 0 to 1 (Graf, 2016). For this reason, the ranges over which the model can predict PID gains are bounded in accordance to each of the three gains. These ranges will be further specified in the next chapter.

**Steps and Episodes**

An episode is defined as starting in the home state, taking an action to adjust the PID gains and then executing the full apple picking motion, that is moving towards the new apple coordinate state, grabbing the apple and moving it back to the start point, where the apple is dropped in the basket. There are no intermediate steps or states in this environment, because the agent returns to the basket after every executed action and the only relevant states are the apple locations. This also means that there is no delay for receiving the reward. Actor-critic methods that only operate from one state at a time are also called zero-step actor-critic. Because steps and episodes have the same meaning in context of this project, they can be used interchangeably. However, for clarity we will refer only to steps from this point on.

**Reward**

The goal of the agent is to choose PID gains that apply a correction to the input signal in such a way that the error between commanded and actual position while moving is minimised. For the reward, this is reflected by the total trajectory error, which is the summed total of absolute position errors for the whole trajectory to the apple coordinate and back. For both trajectories to and from the apple coordinate, the commanded and actual positions of each time step were stored in a log file. After finishing the trajectory, the error could be calculated by subtracting the commanded position from the actual one for each time step, the results of which were stored in a list. These values were made absolute to make sure that negative and positive values would not cancel each other out while summing them. Furthermore, because some trajectories contain more points than others, a spline interpolation function was used to interpolate the errors. By calculating the error over the spline, the amount of points would not impact the total error value. The function used was provided by *scipy*[4]. From the spline, the integral error could then be calculated by calling the scipy function's integral method, resulting in a value that reflects the area of the error. The negative of this integral error forms the reward; in this way a smaller error value forms a better reward than a greater error value. The above elements are formalised in equation 4.1.

$$r = -\int_0^\infty \text{total trajectory error} \qquad (4.1)$$

## 4.3 Advantage Actor Critic

**Advantage Function and Temporal Difference**

In chapter 2, the general concepts behind actor-critic reinforcement learning were explained. It was mentioned that the actor-critic family falls under category of policy gradients, but with the added benefit of a value function that approximates reward for every step: the critic. The critic network made it possibles to use bootstrapping, a technique that reduces variance. It was also mentioned that a popular choice for the critic's value function is the advantage function $A(s, a)$. The advantage function incorporates both the value function $V$ and the Q value function $Q$. The benefit of using the advantage function over just the value or Q value function, is that it subtracts a baseline function from $Q$ in the form of $V$. In the formal definition of the advantage function, which is given by

$$A(s, a) = V(s) - Q(s, a),$$

it can be seen that the advantage function uses the value function $V$ as baseline for $Q$. Because $Q(s, a)$ takes into account the value of a particular action in a state, but $V(s)$ only takes into account the state, the advantage function can be thought of as how valuable a particular action is in a certain state when compared to that state in general. For instance, if a Q value function were to be used instead of the advantage function and the agent would be faced with a state in which all actions lead to negative rewards, then it would also not receive a good value, even if it chose the best possible action for that state. However, with the advantage function, the value of the state-action pair $Q$ would be compared to the value of the state in general $V$ and the agent's critic would attribute a value that reflects whether it chose the best action in that state, even if that action resulted in a negative reward. Because of the use of baseline, the gradients become smaller and the variance of the updates is also reduced. This makes the use of the advantage function in actor critic, also called Advantage Actor Critic or A2C, a popular choice.

---

[4]Scipy Interpolate Univariate Spline: https://docs.scipy.org/doc/scipy/reference/generated/scipy.interpolate.UnivariateSpline.html

It must however be noted that the advantage function has to approximate both the value function and the Q value function in order to compute the advantage. This would in theory lead to having two neural networks for the critic, one for the value function and one for the Q value function. This could be inefficient, but in practise the advantage function can approximated by the temporal difference error

$$A(s_t, a_t) = r_{t+1} + \gamma V(s_{t+1}) - V(s_t) \tag{4.2}$$

The TD error, which is oftentimes denoted by $\delta$, is an unbiased estimator of the advantage function. Using it eliminates the need for two value function estimators, as the critic can instead approximate just the parameterised value function $V$. This method is sometimes referred to as TD actor-critic, although advantage actor critic is also still widely used (Sutton & Barto, 2011). Should the reader be interested, a slightly more detailed derivation of the above equation can be found in appendix section A.3.

**Update Rules**

We can now define the update rules for our advantage actor-critic algorithm. For the actor, we parameterise the policy as $\pi^\theta$, with parameter vector $\theta$, while for the critic, we parameterise the advantage value as $A^w$, with parameter vector $w$. In section 2.2.2, the update rule for the policy parameter $\theta$ was given by

$$\theta \leftarrow \theta + \alpha \nabla_\theta J(\theta) \tag{4.3}$$

With the advantage function defined, we can incorporate it into the object gradient, which can then be defined as

$$\nabla_\theta J(\theta) = \mathbb{E}\left[\left(\sum_t \nabla_\theta \log \pi_\theta(a_t \mid s_t) A_w(s_t, a_t)\right)\right] \tag{4.4}$$

Then, using Monte Carlo approximation to eliminate the expectation from the objective gradient, we can write the update rule for the actor as

$$\theta \leftarrow \theta + \alpha \cdot \nabla_\theta \log \pi_\theta(a|s) A_w(s,a) \tag{4.5}$$

and the critic, the update rule is given by

$$w \leftarrow w + \beta \cdot \delta \cdot \nabla_w A(s,w) \tag{4.6}$$

In these equations, $\alpha$ and $\beta$ denote the learning rates for the actor and critic respectively.

## 4.4 Implementation of A2C on Robot Arm

Having defined the reinforcement learning environment and the mathematics needed for the A2C algorithm, we can describe the implementation of the algorithm on the robot and environment[5]. First, the robot agent receives a set of $[x, y, z]$ coordinates, at which an apple is located that it must pick. Next, it takes an action in the form of adjusting the PID gains of its actuators. This action is decided by sampling from stochastic parameterised policy, which is estimated by the actor neural network. This and the critic network are both built using *PyTorch*[6]. The policy is assumed to be distributed normally, as is the case for most actor-critic implementations and is therefore defined in terms of a mean and standard deviation (Sutton & Barto, 2011). Using the mean and standard deviation, the policy distribution is created from which an action probability can be sampled. This probability can take on a range of values, which might not always be suitable for each PID gain value. For example, as mentioned earlier we want to avoid negative PID gains for system stability and also apply boundaries for each individual gain. This is why a sigmoid function is applied to the action probability sampled from the policy. The sigmoid function ensures the value lies between 0 and 1 and that value can then be multiplied with a scalar for different PID gains. For example, for a gain such as $K_p$, sigmoid function can be scaled with a value of 1000, to ensure that the gain falls between a range of 0 and 1000. This method is applied for each of the different PID gains, $K_p$, $K_i$ and $K_d$. This means that the network outputs a mean and standard deviation for each of these gains, from which an action can then be sampled, which is then scaled in accordance to a range specified for each gain.

Now that the robot agent has decided on an action, the action can be executed. The PID gains of the actuators are set and the robot carries out the movement to the specified apple coordinate. There it picks the apple, moves back to the home state and drops the apple in the basket. The reward is then calculated as defined in above and is based on how accurately the movement was executed. The agent then adjusts both the actor and critic network weights based on this reward. Because the critic update uses the temporal difference error and the actor update uses the critic's value function, the first step is to compute the TD error. Since the initial goal of this project is to test changing the PID gains for one whole movement back and forth, there is only the state of the desired apple coordinates and not a next state before the end of the episode. Therefore, the next state term in the TD error equation is left out and the TD error is simply the reward minus the value of the apple coordinate state. The critic value is then updated using the update rule as defined in equation 4.6. It should be noted that since we use the TD error $\delta$ as estimator for the advantage function $A$, we can substitute $\delta$ for $A$, causing the critic update rule to essentially become the squared TD error. Lastly, the actor weights are updated, as per the rule defined in equation 4.5. An overview of the algorithm is defined in **??**.

---

[5]Code was adapted from: https://github.com/philtabor/Youtube-Code-Repository/blob/master/ReinforcementLearning/PolicyGradient/actor_critic/actor_critic_continuous.py

[6]PyTorch: https://pytorch.org/

**Algorithm 1** Zero-step Advantage Actor-Critic
---
    **for** episode in number of training episodes **do**
        Receive apple coordinate as input state $s$
        Sample action $a$ - the PID parameters- from policy $\pi_\theta(a \mid s)$
        Apply action $a$, execute full motion to apple and back and observe the reward $r$
        Update TD error $\delta \leftarrow r - V(s)$
        Update critic parameters $w \leftarrow w + \beta \cdot \delta \cdot \nabla_w A(s,a)$
        Update actor parameters $\theta \leftarrow \theta + \alpha \cdot \nabla_\theta \log \pi_\theta(a|s) A_w(s,a)$
    **end for**

# Chapter 5:  Experiments

This chapter continues from the general method and defines the specific details of the implementation, as well as how experiments were conducted. First, the types of experiments that will be done will be outlined. Secondly, details concerning the experiment setup, such as network parameters and evaluation types will be discussed. Finally, the last section is entirely dedicated to the different types of errors and crashes that could occur and how these were circumvented.

## 5.1  Experiment Setup

The aim of this research was twofold. First, it was to predict optimal PID gains for one particular apple picking motion and secondly, it was to predict PID gains that were adapted based on where an apple was located. The experiments were therefore separated into these two main categories, which will be referred to as single apple experiments and multiple apple experiments.

### 5.1.1  Single Apple Experiments

Single apple experiments were conducted on one apple coordinate, which was [0.0, 0.625, 0.5] and were trained for 1000 steps. Initially, the first few training rounds for the single apple experiments used only one actuator. This was done in order to test the performance of the algorithm. If it was not able to find good PID gains for one actuator, it probably also would not be able to do this for multiple actuators at the same time and the implementation had to be adjusted. Therefore initial tests were run for only one actuator, while setting all the others to their default gains. Later on a second actuator was also added to test performance of simultaneous PID predicting on multiple actuators.

The actuators that were chosen for training were the shoulder and elbow joint, J1 and J2 respectively, which are both supplied by HEBI Robotics. These joints both had to perform relatively more motions than for example the wrist joint. It was therefore assumed that the two HEBI joints would have a bigger impact on the total motion performance and that tuning their PID controllers would lead to more visible results. The HEBI actuators were also supplied with default gains determined by HEBI Robotics to be well performing. This made it easier to evaluate the actor-critic algorithm: if the gains found by the algorithm outperformed those given by HEBI, then the experiment would be a success.

### 5.1.2  Multiple Apple Experiments

The multiple apple experiments also used the J1 and J2 actuators, but unlike the single apple experiments, no tests were run for individual actuators and J1 and J2 were immediately used together. Furthermore, because training and testing for these experiments had to be conducted on multiple apple coordinates, a training and testing dataset had to be generated. Apple coordinates were generated in accordance to the robot arm being able to reach them. For the x axis, the robot could safely reach 0.5 meters from the rails on both sides, so the x coordinate of an apple could range from -0.5 to 0.5. Consequently, for the y range, the robot only had to be able to grasp apples in front of it, where it could reach a range of 0.5 to 0.75 meters. Finally, for the z-axis, the robot could traverse the rails within a range of 0.3 to 1.5, so the length of the rails without risking touching the ground. However, because movement in the z dimension made the problem more complex, it was set to a constant value, 0.5 meters, for training and testing. This was done to ensure the model could first find good gains for a simpler movement.

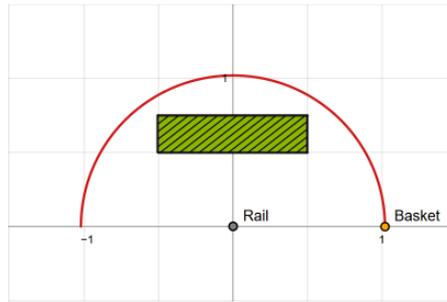

Figure 5.1: In red: the full range of the robot arm. In green: the area where apple coordinates could be sampled from

Apple coordinates were thus generated by sampling from these workspace boundaries. First, a random decimal within range for x, y and z would be generated, creating a coordinate. Then, the Cartesian distance of this coordinate with respect to the previously generated coordinate would be calculated. If this distance fell under a set threshold, in this case 0.01 meters, it would be discarded and a new coordinate would be generated. By doing so it was assured that the model would be trained on a wide set of coordinates. This reduced the chances of the algorithm not being able to predict different PID values for coordinates that were too similar.

For both training and testing 100 different apple coordinates were generated. The algorithm was trained 100 times on these, so one epoch consisted of 100 movements to different coordinates and each epoch was in turn also run 100 times. For each of these epochs, the order of the coordinates was shuffled.

Furthermore, because the multiple apple experiments had to learn how to predict gains for more than one coordinate,

### 5.1.3 General Setup

**Movement between Coordinates**

In both types of experiments, the movement the robot arm had to execute was defined in terms of poses, or coordinates where the end effector, which is the gripper, should be. This meant that a coordinate of an apple could be specified and the arm would move towards that point. Furthermore, for testing purposes, the gripping motion was excluded from the full motion. This was done because the gripping motion was not being optimised by adjusting just the elbow and shoulder joint controllers and leaving the gripping motion out would speed up the learning process. Instead, a pose goal for the apple coordinate and for the home state were specified, where the robot arm would move to and then back from.

**Network Parameters**

The neural networks of both the actor and critic were built using PyTorch[1] and used the Adam optimiser[2]. The first input layer for both networks consisted of one node which takes a vector of three values, the $x$, $y$ and $z$ coordinates of an apple coordinate. Both also had two hidden layers, which each consisted of 256 nodes. The output layer of the critic consisted of only one node, which outputs the value of the actor's chosen action, while the actor's output layer consists of the mean and squared deviation of the amount of PID gains that the mode has to predict. For example, for the single actuator experiments, the algorithm had to predict three PID gains $K_p, K_i$ and $K_d$. The neural network's output dimension would therefore be six, as a mean and standard deviation would be computed for each of the three gains. On the other hand, for experiments with two actuators, the output dimension would be twelve.

---

[1] PyTorch: https://pytorch.org/
[2] Adam Optimiser: https://pytorch.org/docs/stable/optim.html

|  | Actor Output | Critic Output | Epochs | Steps | $\alpha$ | $\beta$ |
|---|---|---|---|---|---|---|
| **Single Apple** Single Actuator | 6 | 1 | 1 | 1000 | 0.0005 | 0.0001 |
| **Single Apple** Two Actuators | 12 | 1 | 1 | 3000 | 0.00005 | 0.00001 |
| **Multiple Apples** Two Actuators | 12 | 1 | 100 | 100 | 0.00005 | 0.00001 |

Table 5.1: Summary of network parameters

Furthermore, a range of learning rates was tested for each setup, beginning with the default learning rate of the Adam optimiser, which is 0.001. For one actuator, it was found that this default value made the model converge too fast, causing it to not be able to find good PID gains. Ultimately the values of 0.0005 for the actor learning rate $\alpha$ and 0.0001 for the critic learning rate $\beta$ were used. These values made sure the model converged within 1000 steps, but not too quickly. For the experiments with two actuators, both single apple and multiple apple, the learning rate needed to be set even lower. These were therefore set to 0.00005 and 0.00001 for actor and critic respectively. It was also suspected that due to added complexity of multiple actuators or apples, the model would require more steps in order to converge. Therefore, for single apple experiments with both actuators, the amount of steps was increased to 3000. For the multiple apple experiments, as was also mentioned above, the algorithm would train in 100 epochs each containing 100 different apple coordinates or steps. All parameters as defined above are summarised in table 5.1.

### Evaluation Methods

The two experiments differ in that they train on either one or multiple apple coordinates. In the case of on apple experiments, this meant that is was not possible to generate training and testing data sets; there was only one data point after all. For this reason, evaluation of the single apple experiments was done by comparing to a baseline performance. As mentioned before, the two actuators were both supplied with default gains, tuned by the HEBI Robotics team, which were found to be well performing. For the shoulder joint J1, these gains were [15, 0, 0] for $K_p$, $K_i$ and $K_d$ respectively and for the elbow joint they were [30, 0, 1]. Initial tests showed that the $K_d$ value for both of these was not yet good performing however, so the gains were further improved by using the ZN-method of tuning PID parameters. The final baseline gains that were used are defined in figure 5.2 below.

|  | Kp | Ki | Kd |
|---|---|---|---|
| **J1** | 15 | 0 | 1 |
| **J2** | 30 | 0 | 1 |

Table 5.2: Baseline PID gains for J1 and J2

By executing the motion multiple times with these gains, an average reward could be specified.

The rewards achieved trough the A2C model could then be compared to this baseline and if they performed better, that would mean the model's found gains were more optimal.

Furthermore, for the multiple apple experiments a training and testing dataset was generated trough the apple sampling technique defined earlier. The model would be trained on the training set, before being applied to the testing set, a set of coordinates the model had never seen before. If performance for the testing set was similar to training, it would mean that the model was successfully able to predict PID gains based on apple coordinates.

## 5.2 Error Handling

### 5.2.1 Aborted Motions

One error that was frequently encountered during the project was that a motion would not complete the full trajectory. Motions were aborted when they had not arrived at the goal points yet, in which case MoveIt would automatically raise an Invalid Trajectory error and clear the current trajectory and goal. Initial tests showed that increasing the acceleration and velocity limits, thus providing the arm with more leeway, made the aborted motions much more likely to occur, while setting heavier limits did the opposite. Following from these test, the abrupt aborting of motions was suspected to be caused by an instability of the arm. While the limits were set to a low value to reduce this error, it was also suspected that a wrong combination of predicted PID gains could cause instability. For instance, some gains could make the arm overshoot drastically, causing it to spin around the rails without reaching its goal, thus causing it to crash. The motion executed after such a crash would then also be more likely to crash itself, even if its chosen gains were not faulty. Another possible cause for the aborted motions was that a deeper problem was present between the communication of MoveIt and the HEBI Robotics interface. After briefly consulting HEBI about the issue, they made it clear that they had indeed found a software problem in their interface that could cause the error we're seeing. Before this issue could be resolved by HEBI however, a way of handling the error needed to be implemented, so that the project could continue.

While it is likely that wrong gain combinations can cause the motion abort, it was also observed that arbitrary gain combinations could cause the same error to occur, even if a crash had not occurred in the previous few steps. Because multiple physical processes are simulated within gazebo, these crashes might be due to an accumulation of some process error over time. Figure 5.2 shows a plot of a all gain combinations that were predicted in a 1000 step run, where the orange points signify a gain combination that caused a crash. It can be seen that crashes typically occur over the entire possible action space. However, it is not clear if the crashes occur due to some process or faulty gain combinations, although the possibility of the errors being caused by faulty gains is also not necessarily excluded by this data. Because the exact causes of the crashes could not be determined precisely, it was important to implement a general way of handling crashes. This was done in the form of a fail-safe.

**Learning from Crashes**

**Fail-safe**

In order to counter these issues, a method was implemented in which the PID gains were automatically set back to the default gains once an invalid trajectory error was detected. Since the default gains were known to be well performing, they would most likely not produce a crash and would be a safe option to use. After setting the gains to safe values, the robot's poses were reset using the reset world simulation service[3] provided by Gazebo. Finally, two full motions back and forth to the apple coordinates would be executed with these settings. This was done two times to ensure that previous executions would not influence the current movement: a full motion executed with safe gains would under normal circumstances not be likely to produce a crash, but if it followed directly after a crashed step, it could be influenced by the previous step and not produce reliable results. This could even be the case for the consecutive step, which is why two full motions were executed before moving on to the regular training. The reset and two motions with safe gains were not counted as steps and the algorithm did not learn from them.

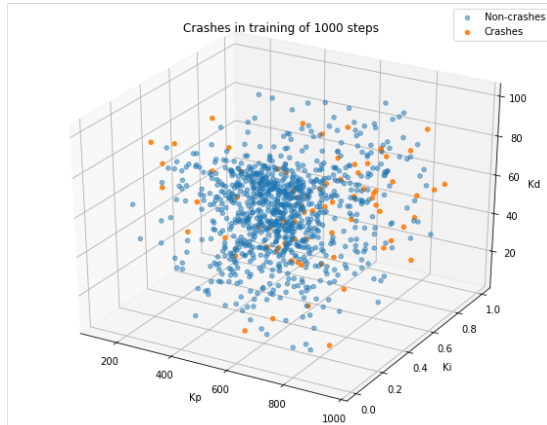

Figure 5.2

In addition to setting the gains and doing the motions, a reward for invalid runs was manually specified, which was set to -3. This value was low enough so that the model would not be able to reach it naturally, but also not so low that it could change model behaviour. For instance, while conducting initial tests, we found that a value of -15 was too drastic and made the model want to avoid that at all costs. It was therefore more likely to choose gains that it considers as safe, instead of exploring better options.

**Motor Overheating**

In addition to the error defined above, the arm would also not be able to perform a motion if its actuators were overheating. The temperature buildup caused by the continuous movement of the robot arm over training could cause the actuators to overheat, which could affect performance. For the HEBI actuators, a builtin function was provided to check if the motors were overheating. This function was used to implement a temporary suspension of the training if one of the HEBI actuators was overheating, allowing the actuator temperature to drop to safe ranges.

### 5.2.2 Exploding Gradients

One final problem that occurred was that the model could start outputting Not a Number values or NaN's instead of valid PID gains. This happened especially if experiments were run for a great amount of steps. The outputting of NaN's was believed to be caused by *exploding gradients*, a problem where gradients increase or decrease at an exponential rate. Exploding gradients often occur in problems with a lot of change. This is why, the problem cause great difficulties for the multi apple experiments, where the model had to learn from multiple input coordinates a a high rate. Well known countermeasures are choice of optimiser, reducing the learning rate, or a technique gradient clipping (Sutton & Barto, 2011). The first two of these options were tested, but did not improve the problem, which is why gradient clipping was implemented. However, the gradient clipping also caused another issue, namely that the aborting of motions mentioned above happened much more frequently. Possible explanations for this might be that computationally heavy gradient clipping was interfering with the HEBI interface, which was already known to contain problems. Because this last issue could only be resolved by HEBI, it became impossible to run long training sessions for the multiple apple experiments. This is why the results below only contain 9 epochs of 100 steps for these experiments.

---
[3]Gazebo Reset Function Documentation: http://gazebosim.org/tutorials/?tut=ros_comm

# Chapter 6: Results

This chapter outlines the results of the experiments. First the results of the single apple coordinate experiments will given. These results are further divided into the single actuator experiments and the two actuator experiments. The results of both will be shown, after which the relations between the two will be discussed. The second section contains the results of the multiple apple experiments, which were only run for two actuators at the same time. Unless specified otherwise, all plots use data from which crashes have been removed. This was done to ensure readability of the figures.

## 6.1 Single Apple Coordinate

### 6.1.1 Single Actuator Experiments

Figure 6.1 shows the rewards achieved over steps while training only the shoulder joint J1, in (a), and training only the elbow joint J2, in (b). The red line denotes the average reward that the baseline PID values achieved, which is a mean of 0.04189 for J1 and 0.17995 for J2. For the plot of J1, it can be seen that the model is often able to reach a higher reward than the baseline and achieving high rewards becomes more frequent in later steps. For J2, the baseline lies higher than for J1 and at first, the model mostly performs under the baseline. However, in later steps the baseline is surpassed.

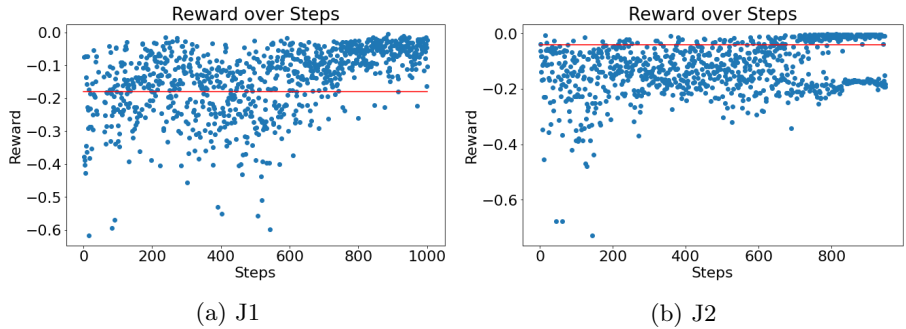

(a) J1  (b) J2

Figure 6.1: Reward over Steps

Figure 6.1 shows the model's predicted PID gains over the training process. For $K_p$, it can be seen that the model converges towards higher values. For $K_i$, at first the model seems to move towards a value of around 0.4, but it seems to distribute again around 1000 steps. Finally, for $K_d$, the model converges towards lower values under 20.

Results of the training done for the J2 actuator are plotted in figure 6.3. In these plots, it can first be observed that $K_p$ also converges towards a high value similarly to J1. However, this convergence is stronger than for J1, as the model narrows to a range 900 towards the end of the steps. For $K_i$, convergence is less clear, although the model does seem to favour lower values towards the end of training. Finally, for $K_d$, the model converges to the boundaries, 0 and 100, at around 700 steps.

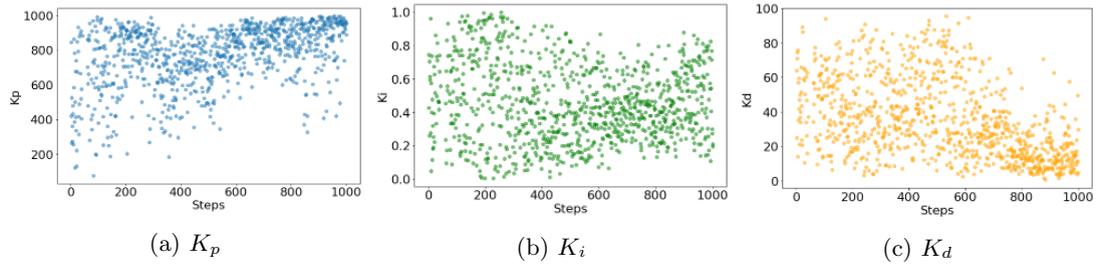

(a) $K_p$        (b) $K_i$        (c) $K_d$

Figure 6.2: Predicted PID values over steps for J1

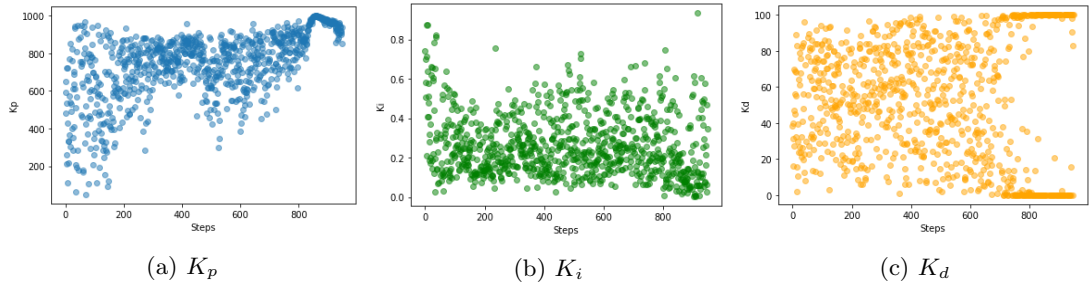

(a) $K_p$        (b) $K_i$        (c) $K_d$

Figure 6.3: Predicted PID values over steps for J2

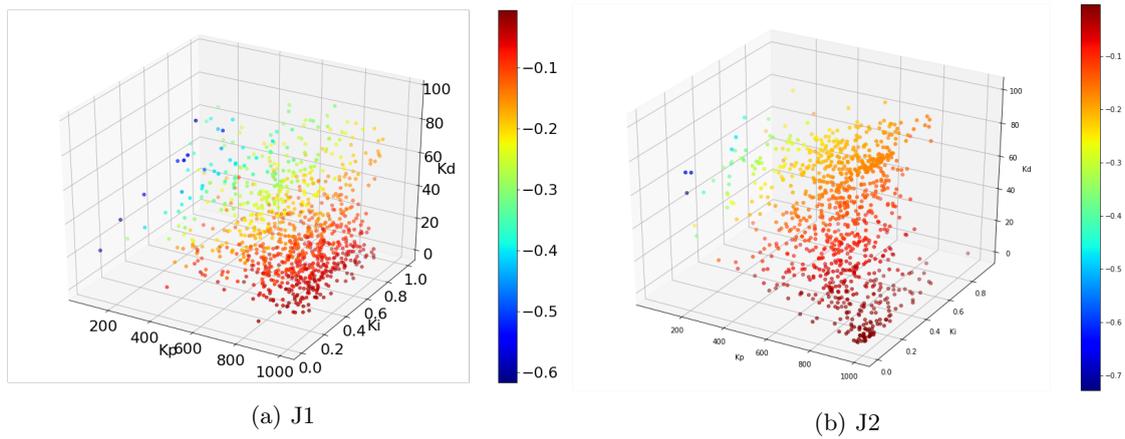

(a) J1        (b) J2

Figure 6.4: Predicted PID gain values over reward

The plot in figure 6.4 show the reward score in colour for all of the three PID gains. Blue signifies a low reward, while dark red signifies a high reward. First, from the reward plot of J1, it can be observed that the highest performing gains are located mostly towards the lower right half. For $K_p$, the best rewards are achieved for higher gains around 800 to 900, which is also what the model was converging to. Furthermore, for $K_i$, higher rewards are spread more evenly over the axis, although the most frequent high rewards occur with a $K_i$ value of around 0.3. Finally, for $K_d$, the best performance is achieved with lower values, ideally under 20, which also corresponds with the converge in figure 6.2.

For J2, the 3D plot shows a slightly different pattern. First, for $K_p$, the best gains lie even higher than J1 and the highest reward are achieved with a $K_p$ value of 900 to 1000. Also, for $K_i$, the most frequent occurrence of high rewards happens with values lower than 0.2. Finally, for $K_d$ lower values still seem to perform the best, although when compared to J1, $K_d$ can perform better with higher values. This was also seen in figure 6.3, where $K_d$ converged to both lower and upper boundaries.

### 6.1.2 Both Actuator Experiments

The following results are for training on both the J1 and J2 actuators simultaneously. Figure 6.5 shows the overall reward achieved by the model over the course of 3000 steps. It can be observed that the reward increases over steps. Furthermore, the red line plotted in denotes the mean performance of the baseline PID gains, which had an average reward of -0.2032. It can be seen that the model is able to reach a higher reward than the baseline and does so more consecutively towards the end of the training.

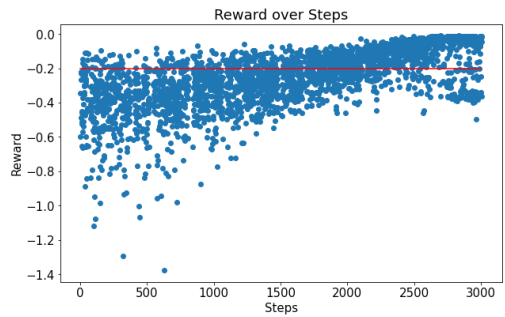

Figure 6.5: Reward over steps for training J1 and J2 simultaneously

The following plots show the results of the single apple experiments with both actuators for actuator J1. In figure 6.6, the values of each of the three PID gains for every step in training are shown. The first plot shows that $K_p$ slowly converges to a relatively high value until around 2000 steps, after which it converges faster to a value of around 800 to 900. For $K_i$ convergence happens slightly more abruptly, with the value converging to a value of around 0.1-0.2 within the last 500 steps. Finally for $K_d$, the value converges to a value that is quite low, almost 0, but it can also be seen that the algorithm starts exploring the upper bound value in the last few steps as well.

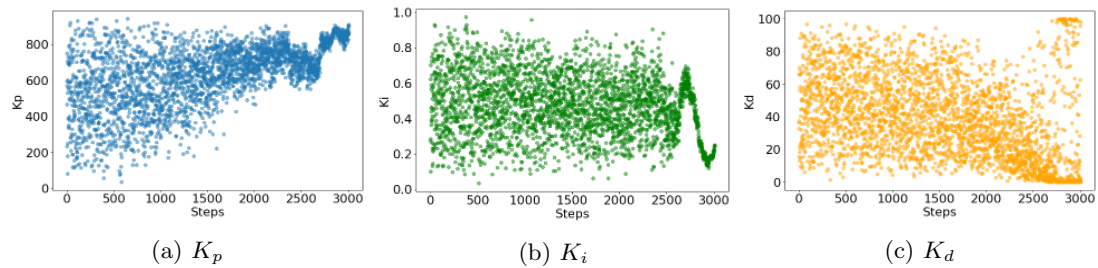

(a) $K_p$  (b) $K_i$  (c) $K_d$

Figure 6.6: Predicted PID values for both actuators - J2

Figure 6.7 shows the predicted PID values with respect to the received reward. It can be observed that the $K_p$ gain is able to achieve a higher reward if its value is also set to a high value; ideally 800 to 900. This is also the range that $K_p$ converged to in the PID value over steps plot. For the $K_i$ gain, the best rewards are generally achieved from a range of 0.2 to 0.6, which also corresponds to the convergence shown in figure 6.2. Finally, for $K_d$, higher rewards are obtained by choosing lower values and a value of around zero has the highest reward. From the 3D plot it can also be derived that all three gains should generally fall into their optimal ranges in order to achieve the highest reward.

The following plot show the results of the single apple experiments with both actuators, for actuator J2. Figure 6.7 shows of the PID gains over training steps. Once again, $K_p$ seems to be converging the most quickly out of the three gains. Similarly to J1, the model converges to a range of about 800 to 900. The $K_i$ gain does not converge as clearly as it did for J1, but it can be observed that the model favours an increasingly lower $K_i$ value over steps, although it also starts to explore the higher values for the last few steps. The third subplot shows an even clearer convergence for $K_d$ than J1, with the value converging to almost zero.

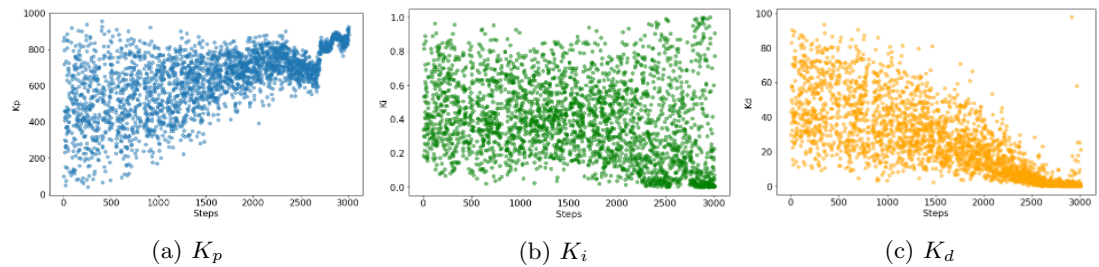

(a) $K_p$      (b) $K_i$      (c) $K_d$

Figure 6.7: Predicted PID values for both actuators - J2

Figure 6.8 again shows the predicted PID values with respect to the reward. For J2, $K_p$ achieves the highest reward with a value of around 600 to 900, which is slightly more broad than the range for J1. $K_i$ is able to achieve a high reward over almost the whole set of values, however, the high reward are much more frequent when $K_i$ has a lower value, ideally around zero. Furthermore, $K_d$ also performs best with a value that is close to zero. It can also be observed that $K_d$ has slightly more influence than with J1, as the highest rewards occur almost exclusively if the $K_d$ value is set higher than 800.

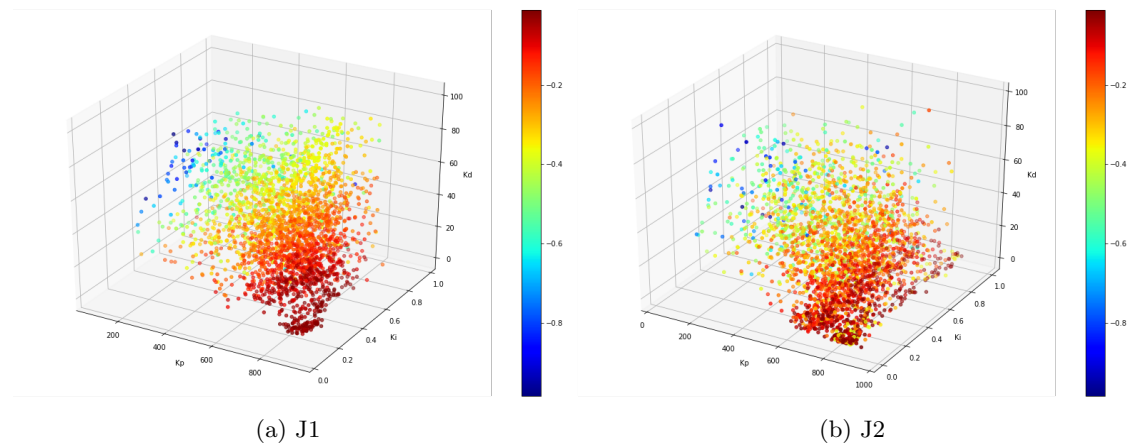

(a) J1      (b) J2

Figure 6.8: Predicted PID gain values over reward

### 6.1.3 Coefficients

To research how the two actuators each influenced the reward and if there was a difference in regards to using one actuator at a time as opposed to two actuators at the same time, a multivariate linear regression was applied to the results of the experiments. From this regression problem the coefficients could be estimated. These coefficients indicate the relationships that variables, in this case the PID gains, might have with a target variable, the total error. The estimated coefficients for both the single actuator experiments and the both actuator experiments are listed in figure 6.1. Note that the total error here is negated, meaning that a positive coefficient indicates a decrease in error, so a positive effect and a negative coefficient indicates an increase of the total error. Figure 6.1 shows four tables: the first column contains the data for the experiments with both actuators training at the same time and the second column contains the experiments on single actuators. Within these column, a division is made between data where crashes have been excluded and data where crashes are note excluded. This was done to make the different influences PID gains might have on crashes more clear.

In figure (a), it can be noted that the variable with the most positive effect on the total error is $K_i$ for actuator J1. Also noteworthy is that all gains for J1 have almost double the coefficient value of the J2 counterparts. In comparison, for the J1 and J2 actuators that were run separately from each other, shown in figure (b), the most positive influence on the error comes from $K_i$ for J2, although there are less clear outliers than for figure (a). For figure (c) and (d) which also include crashes, the difference between simultaneous and independent training actuators becomes more apparent. The most positive influence for both actuators (c) comes from J1 $K_p$, although this influence is not very high and for the independent actuators (d) the most positive influence is J2 $K_i$. This last variable is however the most negative influence for the total error of both actuator tuning.

|        | Total Error |
|--------|-------------|
| J1 $K_p$ | 0.000246 |
| J1 $K_i$ | 0.161837 |
| J1 $K_d$ | -0.004195 |
| J2 $K_p$ | 0.000160 |
| J2 $K_i$ | 0.054628 |
| J2 $K_d$ | -0.002511 |

(a) J1 and J2 used simultaneously, crashes excluded.

|        | Total Error |
|--------|-------------|
| J1 $K_p$ | 0.000162 |
| J1 $K_i$ | 0.018437 |
| J1 $K_d$ | -0.003177 |
| J2 $K_p$ | 0.000235 |
| J2 $K_i$ | 0.074919 |
| J2 $K_d$ | -0.002065 |

(b) J1 and J2 used independently, crashes excluded.

|        | Total Error |
|--------|-------------|
| J1 $K_p$ | 0.000357 |
| J1 $K_i$ | -0.089056 |
| J1 $K_d$ | -0.004412 |
| J2 $K_p$ | 0.000185 |
| J2 $K_i$ | -0.121808 |
| J2 $K_d$ | -0.006347 |

(c) J1 and J2 used simultaneously, crashes included.

|        | Total Error |
|--------|-------------|
| J1 $K_p$ | 0.001609 |
| J1 $K_i$ | -0.120123 |
| J1 $K_d$ | -0.002428 |
| J2 $K_p$ | 0.000773 |
| J2 $K_i$ | 0.130234 |
| J2 $K_d$ | -0.003016 |

(d) J1 and J2 used independently, crashes included.

Table 6.1: Coefficients of training for one or two actuators

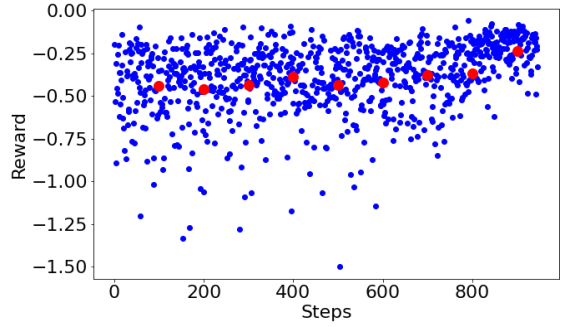

Figure 6.9: Reward over Steps. Baseline in red.

## 6.2 Multiple Apple Coordinates

### 6.2.1 Training

The following results are from the training experiments where the model had to predict PID gains based on apple coordinates. As mentioned before, due to repeated errors these experiments could not be completed, which is why the available data is only for 9 epochs of 100 steps. However, the reward over these epochs as plotted in figure 6.9, shows that there was a visible increase in reward over training. The achieved rewards, shown in blue, can be compared to the average reward of the baseline gains for every epoch in red. This baseline had a reward mean of -0.2215 and a standard deviation of -0.16191. Like in the single apple experiments, it can be seen that the model is able to get a higher reward than the baseline. The last 100 steps also show an increase in model reward, with points more frequently performing higher than the baseline.

Figure 6.10 also compares the baseline performance to the model performance. In the plots, the reward for the different sampled apple coordinates can be seen. Figure 6.10(a) shows the performance of the baseline PID values on every coordinate. The average reward of the baseline -0.22154, with a standard deviation of -0.16191. For the model performance, as shown in figure 6.10(b), the reward around the same value. However, 66 apple coordinates were reached with a higher reward than the baseline. Also noteworthy is that coordinates on the left side of both plots generally achieve lower rewards. These coordinates are the ones that are further removed from the home position.

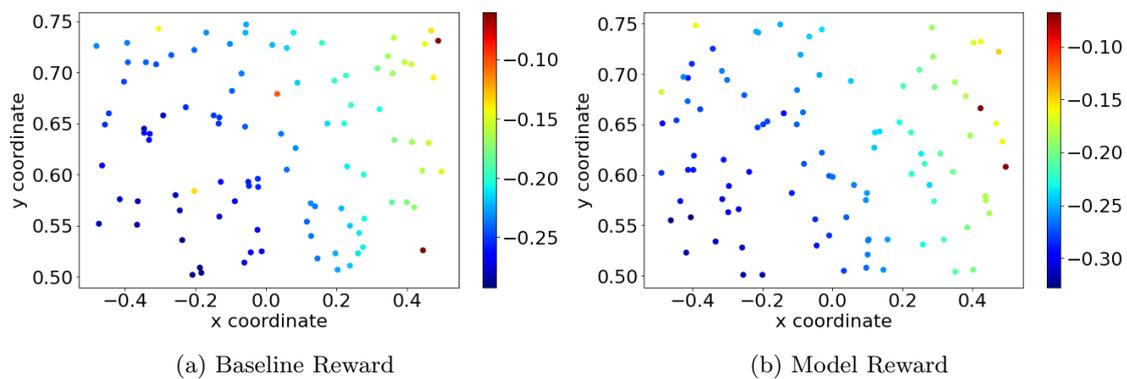

(a) Baseline Reward          (b) Model Reward

Figure 6.10: Reward for coordinates in training set trough use of baseline and model

Figure 6.11 shows the plotted PID gains over the course of the training step for J1s. It can be seen that $K_p$ converges within the 900 steps, reaching towards the upper boundary 1000. Furthermore, $K_i$ also seems to be moving towards convergence, as the upper and lower boundaries 1 and 0 become more frequent over steps. Consequently, $K_d$ can also seen to be moving towards a specific range of values, in this case its lower boundary zero.

Figure 6.12 shows the same PID gains over steps, but for J2. Firstly, for $K_p$ a similar pattern to J1 can be observed, where the value goes towards the upper boundary around 800 steps. For $K_i$ however, convergence is not so apparent. The values seem to move towards 0.6-0.8, but further training would be needed to see if the model converged. Furthermore, for $K_d$, a slight move towards convergence can be observed around the last few steps, where the values range from 60 to 80, but for $K_d$ as well convergence is not completely clear.

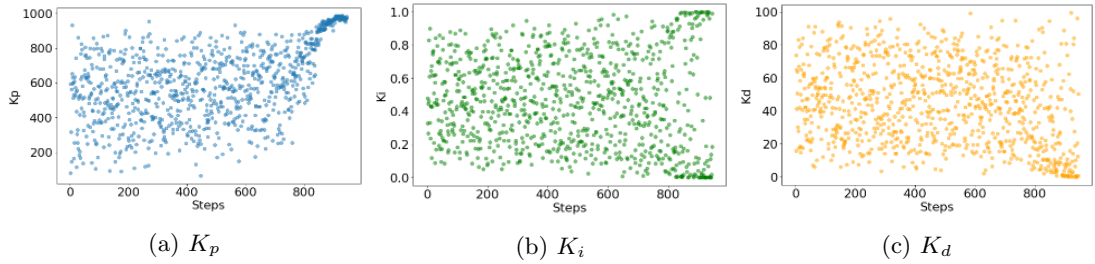

(a) $K_p$  (b) $K_i$  (c) $K_d$

Figure 6.11: Predicted PID gains over steps for multi apple experiments J1

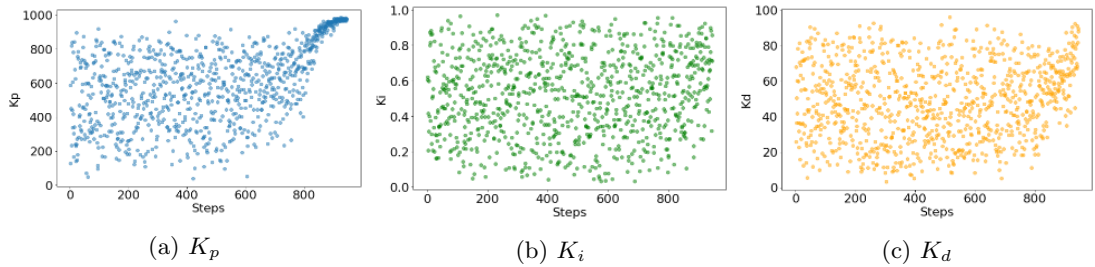

(a) $K_p$  (b) $K_i$  (c) $K_d$

Figure 6.12: Predicted PID gains over steps for multi apple experiments J2

Finally, figure 6.13 shows the PID gains and their rewards for both J1 in (a) and J2 in (b). Although the rewards generally lie somewhat lower than those of the single apple experiments, it can still be observed that a high $K_p$ value is important in regards to achieving high rewards. Furthermore, $K_i$ performs best with a value around 0 or 1 and $K_d$ is better kept under a value of 20. For J2, $K_p$ also seems to perform better from a range of 800-1000. Consequently, reward for $K_i$ increases slightly with a value of 1. Lastly, $K_d$ has the highest reward for a range of 60-80, which, like all other gains, corresponds to the training data shown in figure 6.11 and 6.12.

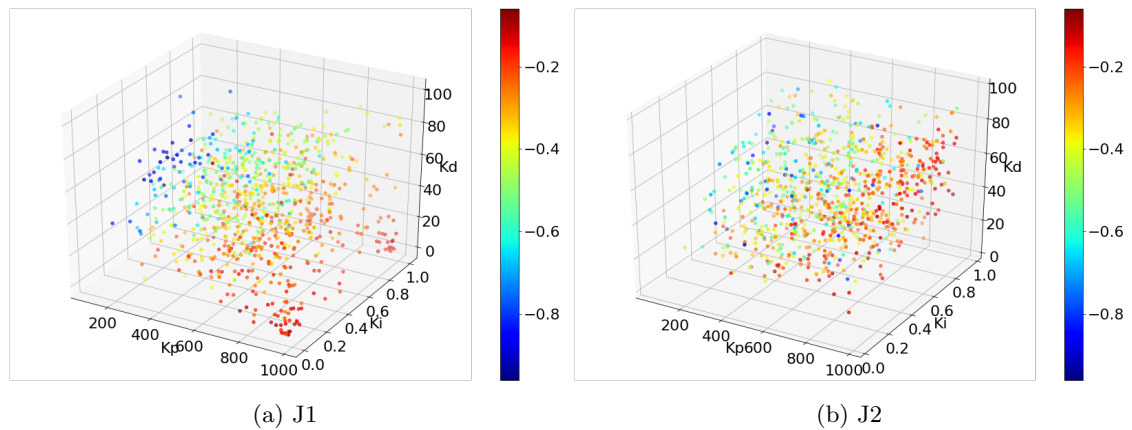

(a) J1  (b) J2

Figure 6.13: Predicted PID gains and their rewards

## 6.2.2 Testing

The trained model was also to a set of testing data. Figure 6.14 shows the rewards over the coordinates for the baseline gains and the model gains. A total of 95 apple coordinates were tested and compared to the baseline and the model improved over 75 of these coordinates. Even coordinates that had low reward for the baseline, such as those in the left part of the plot, achieved a relatively high reward when using the model PID gains.

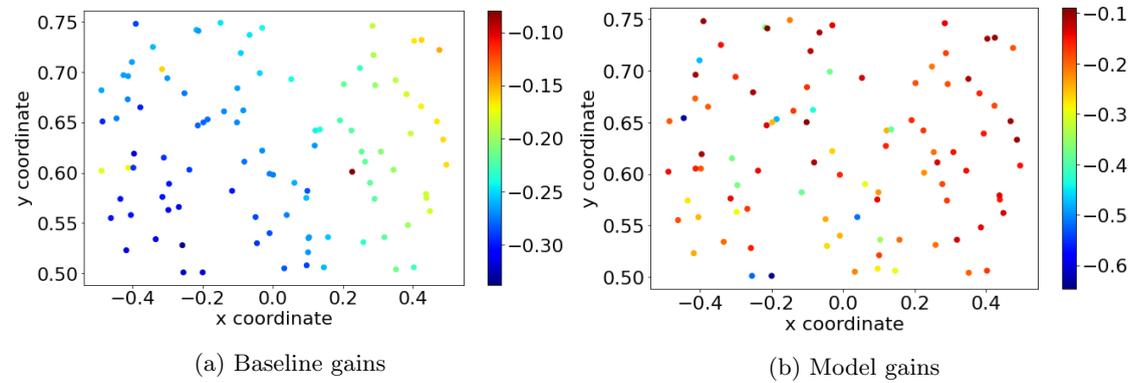

(a) Baseline gains  (b) Model gains

Figure 6.14

# Chapter 7: Evaluation

In this final chapter, the results will first be evaluated and discussed. After the discussion of the results, the conclusions of this project will be given. Finally, the chapter will conclude with a future works section, in which the possibilities for further research will be discussed.

## 7.1 Discussion of Results

For all experiments, the reward as compared to the baseline was plotted. These results all showed that the model could achieve a higher reward than the baseline gains. In addition, these higher rewards became more frequent over time, suggesting that the model was indeed learning. Many of the experiments also showed convergence and comparing these convergences to the rewards showed that the model always converged towards a values that resulted in higher rewards. For $K_p$, the converged value typically lied in a higher range, most of the time ranging from 800 to 1000. This implies that the robotic arm performs better with a higher $K_p$ value. Furthermore, for $K_i$, the overall range of convergence was much more distributed. Sometimes the reward seemed to converge to the boundaries, but sometimes this was also not the case. An optimal value for $K_i$ would therefore require more experiments. Thirdly, $K_d$ converged towards zero most of the time and reward plots also showed that a low $K_d$ performed the best in terms of reward. This suggests that $K_d$ can best be set to a low value. These findings are in line with PID control theory, as $K_p$ is often seen as the most important of the three gains and $K_i$ and $K_d$ often causing more instability.

Secondly, the differences in performance of shoulder joint actuator J1 and elbow joint actuator J2 were compared. Between single actuator experiments for J1 and J2, the reward plots showed that J1 generally had more room to improve, with a baseline that was lower than J2. It also did not converge as fast as J2, which suggests that the gains for J1 were generally harder to learn. For training the two actuators simultaneously however, both J1 and J2 converged at around the same step, although for J2 the $K_i$ value did not clearly converge. This implies that there is a difference in tuning the two actuators independently, as opposed to training them at the same time. The coefficients calculated for these experiments also reflect this. There are great differences between the single actuator and both actuator coefficients, for example with $K_i$ for J1 being an important factor when training two actuators, but not for a single actuator, or $K_i$ for J2 being a negative factor on crashes for two actuators, while being a positive influence on crashes for a single actuator. Overall, tuning both actuators at the same time showed the most promising results: although it required slower learning, optimal gains were ultimately chosen more frequently if both actuators were controller by the model simultaneously.

Thirdly, the results for the multiple apple experiments, although limited, showed that the model was capable of improving over the baseline. Convergence could also be observed in section 6.2.1, especially for $K_d$. The values that were being converged to also showed similarities to the other experiments, with the only exclusion being the $K_d$ value of J2, which converged to a higher value as opposed to a lower one. However hard claims about his should only be made once the experiments are able to run for a greater amount of time.

One of the most noteworthy results is the testing of the multiple apple model on the testing. As opposed to the plots shown in figure 6.10, where only a small improvement over the baseline was observed, the comparing of model performance on testing data showed a great increase in performance. The reward over coordinate plots in figure 6.10 and 6.14 also show that the reward increases over the x-axis. The further to the right an apple coordinate is located, the more likely both baseline and model gains become to achieve a high reward. The home position, where the harvesting basket is located, is also located in the far right lower corner of the plots. This means that the reward decreases as the distance from home -and thus also the amount of required motion- increases. In figure 6.14(b) however, it can be seen that the model performs better even on these far lying coordinates that are otherwise difficult to reach accurately. The model achieving high rewards for these coordinates suggests that it is able to adapt its PID predictions for these points.

## 7.2 Conclusion

The goal of this project was conduct an exploratory research into the possibility of applying deep reinforcement learning to tune the PID parameters of a robotic arm. This robot was simulated in a virtual environment to ensure that multiple research directions could be explored safely. Furthermore, the choice was made to use an A2C algorithm, which combines both policy-based and value-based reinforcement learning techniques. The policy-based policy gradient made it possible to predict PID gains in the form of a continuous range of numbers, while the value-based temporal difference made the algorithm more easily expandable to complex multi-step environments for future research.

In order to study the performance of the A2C PID controller, two main experiments were conducted: one where the model was trained on a single apple coordinate and one where the model had to adapt its PID predictions to a ranging set of apple coordinates. Results of these experiments showed that the A2C PID controller could find PID gains that were better than the baseline for all experiments. Moreover, $K_p$ and $K_d$ in specific were found to be essential gains that had clear ranges of optimal values. For $K_p$, this was a high value of around 800 to 1000 and for $K_d$ a much lower value of around 0 to 20. Gain combinations that had both $K_p$ and $K_d$ set to these ranges were almost exclusively found to be better performing than the baseline, meaning they were more accurate. These results were also compared for tuning one actuator as opposed to the model training two actuators simultaneously. Comparisons suggested that overall performance was improved if the model had control over both of the actuators.

Furthermore, the experiments on multiple apple coordinates showed that a relation exists between apple position and appropriate PID gains. Motions towards coordinates that were located far from the home state were found to be much less accurately executed. Initial results of applying the model to a testing dataset were very positive, as the accuracy for almost all apple coordinates was improved over the baseline, even for motions to coordinates that were harder to reach.

## 7.3 Future Research

One of the first directions for future works could be to run the multiple apple coordinate experiments more exhaustively. Although initial results are promising, the model would need to train for a longer period of time before any hard conclusions might be made. Following this future research might also look into the possibility of making the model more adaptive. PID values could be predicted for multiple steps within the full motion. For instance, a first step could be to make the controller adaptive to apple weight. Other reinforcement learning techniques might also be implemented, such as the more complex A3C algorithm.

# Appendices

# Appendix A: Additional Reinforcement Learning Background

In this section, some more important concepts of both value-based and policy-based methods will be introduced by briefly outlining well-known techniques for each of the methods that were not mentioned in the theoretical background: Q-learning for value-based methods and REINFORCE for policy-based methods.

## A.1 Q-learning

One of the most well known implementations of temporal difference learning is Q-learning. In Q-learning the quality, or Q value, of a certain action in a given state is calculated, which can then be used to determine an optimal policy. The Q value is similar to the value $V$ defined above, but the only difference being that the value $V$ determines the value of a state, while the value $Q$ determines the value of a state and an action. The Q value is therefore defined as the expected future reward that an agent will receive assuming that it is in a certain state $s$, takes action $a$ and continues with the current policy $\pi$ until the end of the episode. The Q function is formalised as

$$Q(s,a) = r(s,a) + \gamma \max_a Q(s',a) \tag{A.1}$$

This formula is also called the Bellman equation (Bellman, 1954). In this equation, the resulting Q value from being in state $s$ and taking action $a$, is the immediate received reward $r(s,a)$ added to the discounted maximum Q value, which can be obtained by taking the action $a'$ in the next state $s'$ that maximises Q. With the Q function defined, we can now define the update rule for Q-learning, using the temporal difference update rule as defined in 2.5

$$Q(s_t, a_t) \leftarrow Q(s_t, a_t) + \alpha \Big(r_t + \gamma \max_a Q(s_{t+1}, a) - Q(s_t, a_t)\Big) \tag{A.2}$$

Using this equation, a Q value can be calculated for every state-action pair the agent encounters in the environment. These Q values are then stored in a matrix, the *Q-table*, which provides a reference for the agent to choose an action such that the Q value of the next state is maximised. By iterating over the environment trough trial and error, the agent can update the Q-table with each step, thus eventually allowing it to determine the optimal values, which in turn can be used to determine the optimal policy.

Of course, if the environment is sufficiently complex and has a very large state or action space, the Q-table can quickly become very large, making it computationally expensive to store and retrieve data from the table. A way to counter this is to use a neural network to approximate the Q value function, instead of computing the value and storing it in a table. This technique is known as *Deep Q-learning* (Mnih et al., 2013). In deep Q-learning, a state is given as input to a neural network, which approximates the Q value function and outputs the Q values for every possible action in the given state. From here, the action corresponding to the highest Q value can be chosen, eliminating the need to store all Q values in a table.

## A.2 REINFORCE

The theoretical background describes the gradient in a general way which is called the vanilla policy gradient (Szepesvári, 2010). Many more policy gradient algorithms exist, one of the most well known being the REINFORCE algorithm (Sutton & Barto, 2011). The REINFORCE algorithm uses stochastic gradient ascent to update the parameter vector $\theta$. It is essentially the same as as vanilla policy gradient, save for a notational difference. The vanilla policy gradient as defined in equation 2.4 contains an expectation in the objective gradient, which can not be directly calculated, as it involves and integral over the policy's probability distribution, which is not explicitly defined. In order to omit the expectation from the gradient, Monte Carlo approximation can be used. Monte Carlo approximation assumes that the integral of an expected value can be approximated by taking the mean over a sufficiently large number of samples (Kalos & Whitlock, 1986). This means that equation 2.4 can be rewritten to the following form, where the expectation has been replaced by a mean over the total samples.

$$\nabla_\theta J(\theta) \approx \frac{1}{N} \sum_{i=1}^{N} \left[ \left( \sum_t \nabla_\theta \log \pi_\theta(a_{i,t} \mid s_{i,t}) \right) \left( \sum_t r(s_{i,t}, a_{i,t}) \right) \right] \quad (A.3)$$

With the rewritten equation, it becomes possible to update the policy's parameter vector by performing gradient ascent. More than often a neural network is used to approximate the policy function, meaning the network's weights make up the parameters vector $\theta$. The network outputs a probability distribution of actions, the most likely action is executed and a reward is subsequently received. The network updates it's parameters only at the end of an episode, according to the total reward received from the trajectory. Ultimately this results in actions with high rewards becoming more likely in the policy's probability distribution and actions with poor rewards becoming increasingly less likely. Algorithm 1 shows the basic steps of the REINFORCE algorithm. (Probably not needed. Or need to do for Q-learning as well)

---
**Algorithm 2** REINFORCE
---
1: Sample trajectories $\{\tau_i\}_{i=1}^{N}$ by running policy $\pi_\theta(a_t \mid s_t)$
2: Set gradient to $\nabla_\theta J(\theta) = \sum_i (\sum_t \nabla_\theta \log \pi_\theta(a_t^i \mid s_t^i))(\sum_t r(s_t^i, a_t^i))$
3: Update using update rule $\theta \leftarrow \theta + \alpha \nabla_\theta J(\theta)$
---

## A.3 TD as estimator of the Advantage

As mentioned before, temporal difference learning is concerned with finding an optimal policy $\pi$ for which the cumulative reward is maximised, by estimating the corresponding value function. This means that the value function that optimises the policy, also called the state value Bellman optimality equation, is defined as follows

$$V^*(s) = \max_\pi V^\pi(s) \quad (A.4)$$

where the optimal value function $V^*$ is the function that maximises the the value function of the policy. Similarly, the Bellman optimality equation can be defined for the state-action value function, or Q value function

$$Q^*(s, a) = \max_\pi Q^\pi(s, a) \quad (A.5)$$

As mentioned in section 2.2, the Q value function denotes the expected future reward that an agent will receive assuming that it is in a certain state, takes an action and continues with the current policy thereafter. Using this and the optimality equations for state value and state-action value, we can write $Q^*$ in terms of $V^*$

$$Q(s_t, a_t) = \mathbb{E}\big[r_{t+1} + \gamma V^*(s_{t+1})\big] \quad (A.6)$$

with which we can in turn define the advantage as

$$A(s_t, a_t) = r_{t+1} + \gamma V(s_{t+1}) - V(s_t) \tag{A.7}$$

Recalling section 2.2.2, it becomes apparent that this is actually the temporal difference error, also known as $\delta$. The TD error is an unbiased estimator of the advantage function. Using it eliminates the need for two value function estimators, as the critic can instead approximate just the parameterised value function $V$.